%% file: learned-syn-idioms.tex
\PassOptionsToPackage{sort&compress, numbers}{natbib}
\PassOptionsToPackage{usenames, dvipsnames, svgnames}{xcolor}
\documentclass{article}
\usepackage{eufrak}
\usepackage{pbox}

\usepackage{polozov_paper}
\usepackage[final]{neurips_2019}

\title{Program Synthesis and Semantic Parsing \\ with Learned Code Idioms}
\author{%
    Richard Shin\thanks{Work done partly during an internship at Microsoft Research. } \\
    UC Berkeley \\
    \email{ricshin@berkeley.edu} \\
    \and\hspace{-13pt}
    \textbf{Miltiadis Allamanis, Marc Brockschmidt \& Oleksandr Polozov} \\
    Microsoft Research \\
    \email{{miallama,mabrocks,polozov}@microsoft.com}
}

\begin{document}
\input{macros}
\maketitle
\setcounter{footnote}{0}

\input{sections/abstract}
\input{sections/introduction}

\input{sections/overview}

\input{sections/mining}

\input{sections/synthesis}
\input{sections/evaluation}
\input{sections/related}
\input{sections/conclusion}
\small
\bibliography{learned-syn-idioms}
\bibliographystyle{abbrvnat}

\newpage
\appendix
\input{sections/appendix-spider-encoder}

\end{document}

%% file: macros.tex
\newcommand{\system}{\textsc{Patois}\xspace}

\newcommand{\fix}[1]{{\color{OrangeRed} \bfseries [AP]:~#1}}

\newcommand{\dataset}{\mathcal{D}\xspace}
\newcommand{\grammar}{\mathcal{G}\xspace}
\renewcommand{\ast}{T\xspace}
\newcommand{\nlseq}{X\xspace}
\newcommand{\word}{x\xspace}
\newcommand{\nlseqemb}{\hat{\nlseq}\xspace}
\newcommand{\wemb}{\vect{\hat{\word}}\xspace}
\newcommand{\hidden}{\vect{h}\xspace}
\newcommand{\tok}{y\xspace}
\newcommand{\linearSeq}{Y\xspace}
\newcommand{\lsTok}{\tok\xspace}
\newcommand{\vocab}{\mathcal{V}\xspace}
\newcommand{\model}{f\xspace}
\newcommand{\action}{a\xspace}
\newcommand{\encoder}{\model_{\text{enc}}}
\newcommand{\decoder}{\model_{\text{dec}}}
\newcommand{\lab}{\ell\xspace}
\newcommand{\idiom}{\mathcal{I}\xspace}
\newcommand{\op}{\mathsf{op}}
\newcommand{\production}{R\xspace}

\lstdefinelanguage{UAST}{
    morekeywords={if},
    otherkeywords={>=,<=,\&\&},
    sensitive=true
}

\newcommand{\circled}[1]{\raisebox{.5pt}{\textcircled{\raisebox{-.8pt}{\small #1}}}}
\newcommand{\holed}[1]{\text{\tcbox[nobeforeafter, boxsep=2pt, box align=base]{#1}}}

%% file: sections/abstract.tex
\begin{abstract}
    Program synthesis of general-purpose source code from natural language specifications is challenging due to the need
    to reason about high-level patterns in the target program and low-level implementation details at the same time.
    In this work, we present \system, a system that allows a neural program synthesizer to explicitly interleave
    high-level and low-level reasoning at every generation step.
    It accomplishes this by automatically mining common \emph{code idioms} from a given corpus, incorporating
    them into the underlying language for neural synthesis, and training a tree-based neural synthesizer to use these
    idioms during code generation.
    We evaluate \system on two complex semantic parsing datasets and show that using learned code idioms
    improves the synthesizer's accuracy.
\end{abstract}

%% file: sections/introduction.tex
\section{Introduction}%
\label{sec:introduction}

Program synthesis is a task of translating an incomplete specification (\eg natural language, input-output examples, or
a combination of the two) into the most likely program that satisfies this specification in
a given language~\citep{gulwani2017survey}.
In the last decade, it has advanced dramatically thanks to the novel neural and neuro-symbolic
techniques~\citep{robustfill,deepcoder,ngds}, first mass-market applications~\citep{flashmeta}, and massive
datasets~\citep{karel,conala,spider}.
\Cref{fig:intro:examples} shows a few examples of typical tasks of program synthesis from natural language.
Most of the successful applications apply program synthesis to manually crafted domain-specific languages (DSLs) such
as FlashFill and Karel, or to subsets of general-purpose functional languages such as SQL and Lisp.
However, scaling program synthesis to real-life programs in a general-purpose language with complex control
flow remains an open challenge.

We conjecture that one of the main current challenges of synthesizing a program is insufficient separation
between high-level and low-level reasoning.
In a typical program generation process, be it a neural model or a symbolic search, the program is
generated in terms of its \emph{syntax tokens}, which represent low-level implementation details of
the latent high-level \emph{patterns} in the program.
In contrast, humans switch between high-level reasoning (\textit{``a binary search over an array''}) and
low-level implementation (``\mbox{\lstinline[language=Python]{while l < r: m = (l+r)/2}\ \dots}'') repeatedly
when writing a single function.
Reasoning over multiple abstraction levels at once complicates the generation task for a model.

\input{sections/fig-examples}

This conjecture is supported by two key observations.
First, recent work~\citep{coarse2fine,bayou} has achieved great results by splitting the synthesis process into
\emph{sketch generation} and \emph{sketch completion}.
The first stage generates a high-level sketch of the target program, and the second stage fills in missing details in
the sketch.
Such separation improves the accuracy of synthesis as compared to an equivalent end-to-end generation.
However, it allows only one stage of high-level reasoning at the root level of the program, whereas
\textbf{(a)} real-life programs involve common patterns at all syntactic levels, and
\textbf{(b)} programmers often interleave high-level and low-level reasoning during implementation.

Second, many successful applications of inductive program synthesis such as FlashFill~\citep{flashfill}
rely on a manually designed DSL to make the underlying search process scalable.
Such DSLs include high-level operators that implement common subroutines in a given domain.
Thus, they
\textbf{(i)} compress the search space, ensuring that every syntactically valid DSL program expresses some
useful task, and
\textbf{(ii)} enable logical reasoning over the domain-specific operator semantics, making the search efficient.
However, DSL design is laborious and requires domain expertise.
Recently, \citet{scc} showed that such DSLs are learnable in the classic domains of inductive program synthesis;
in this work, we target general-purpose code generation, where DSL design is difficult even for experts.

In this work, we present a system, called \system, that equips a program synthesizer with automatically learned
high-level \emph{code idioms} (\ie common program fragments) and trains it to use these idioms in program generation.
While syntactic by definition, code idioms often represent useful semantic concepts.
Moreover, they \emph{compress} and \emph{abstract} the programs by explicitly representing common patterns with
unique tokens, thus simplifying generative process for the synthesis model.

\system has three main components, illustrated in \Cref{fig:intro:overview}.
First, it employs nonparameteric Bayesian inference to mine the code idioms that frequently occur in a given corpus.
Second, it marks the occurrences of these idioms in the training dataset as new named operators in an extended grammar.
Finally, it trains a neural generative model to optionally emit these named idioms instead of the original code
fragments, which allows it to learn idiom usage conditioned on a task specification.
During generation, the model has the ability to emit entire idioms in a single step instead of multiple steps
of program tree nodes comprising the idioms' definitions.
As a result, \system interleaves high-level idioms with low-level tokens at all levels of program
synthesis, generalizing beyond fixed top-level sketch generation.

We evaluate \system on two challenging semantic parsing datasets:
Hearthstone~\citep{hearthstone}, a dataset of small domain-specific Python programs,
and Spider~\citep{spider}, a large dataset of SQL queries over various databases.
We find that equipping the synthesizer with learned idioms improves its accuracy in generating
programs that satisfy the task description.
\input{sections/fig-system}

%% file: sections/fig-examples.tex
\begin{table*}[t]
    \centering
    \caption{Representative program synthesis tasks from real-world semantic parsing datasets.}
    \label{fig:intro:examples}
    \small
    \lstset{basicstyle = \tiny\ttfamily}
    \begin{tabular}{>{\raggedright}p{1.4cm}>{\scriptsize}p{5.9cm}p{5.4cm}}
        \toprule
        \textbf{Dataset} & \small \textbf{Natural Language Specification}  & \textbf{Program} \\
        \midrule
        Hearthstone \citep{hearthstone} &
        \itshape
        Mana Wyrn (1, 3, 1, Minion, Mage, Common) \par Whenever you cast a spell, gain +1 Attack. &
        \vspace{-\baselineskip}
        \begin{lstlisting}[gobble=12, language=Python, belowskip=-1.5\baselineskip]
            # \dots
            def create_minion(self, player):
                return Minion(1, 3, effects=[Effect(SpellCast(), ActionTag(Give(ChangeAttack(1)), SelfSelector()))])
        \end{lstlisting}
        \vspace{-\baselineskip}
        \\
        \midrule
        Spider \citep{spider} &
        \itshape
        For each stadium, how many concerts are there? \par \vspace{5pt}
        Schema: \par \texttt{stadium = \{stadium\_id, name, \dots\}, \dots } &
        \vspace{-\baselineskip}
        \begin{lstlisting}[gobble=12, language=SQL, belowskip=-\baselineskip]
            SELECT T2.name, COUNT(*)
            FROM concert AS T1 JOIN stadium AS T2
                ON T1.stadium_id = T2.stadium_id
            GROUP BY T1.stadium_id
        \end{lstlisting}
        \vspace{-\baselineskip}
        \\
        \bottomrule
    \end{tabular}
    \vspace*{-0.3\baselineskip}
\end{table*}

%% file: sections/fig-system.tex
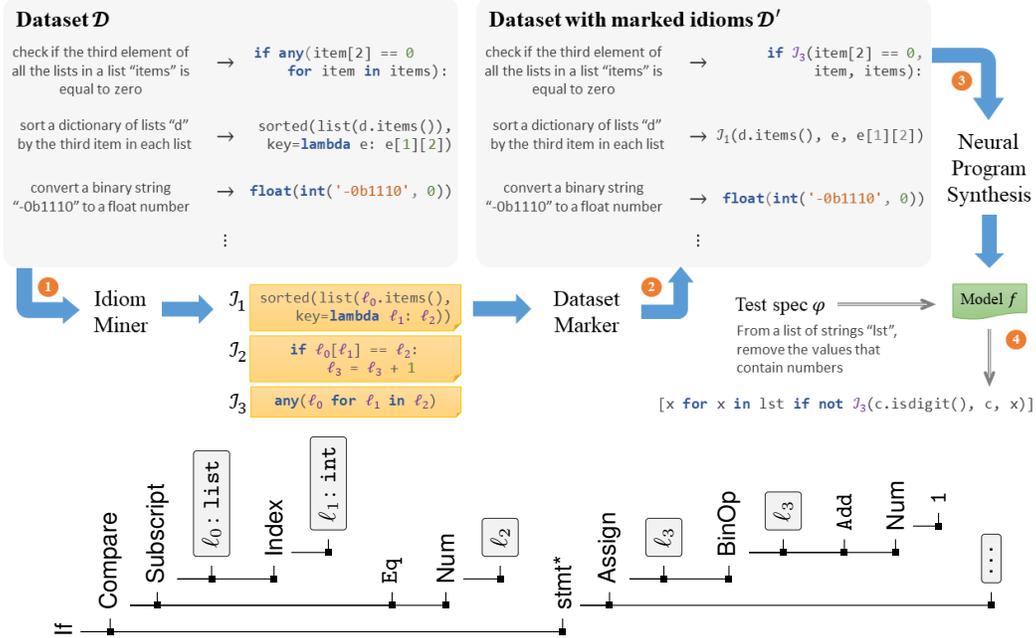
\begin{figure*}[t]
    \begin{center}
        \includegraphics[width=\textwidth, trim=0 155pt 0 0, clip]{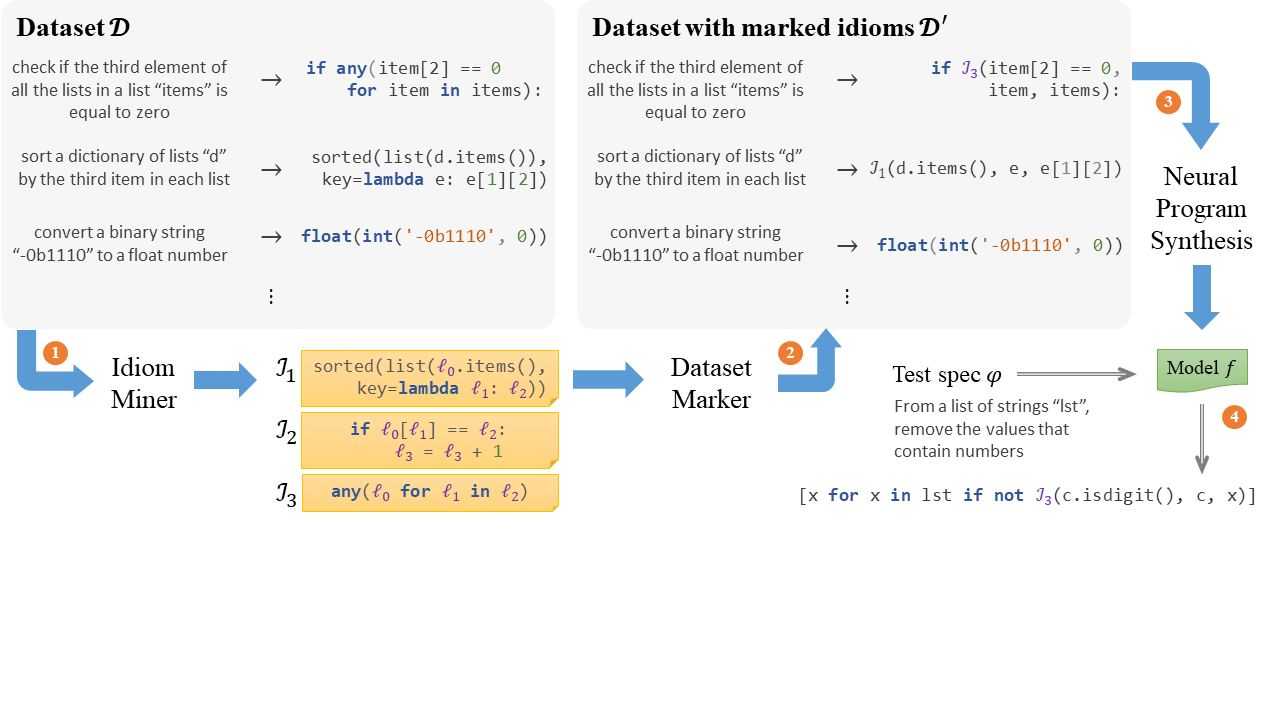}
    \end{center}
    \vspace{-3pt}
    \tcbset{size=fbox}
    \centering
    \begin{sideways}
        \begin{forest}
            for tree={
                font=\small\sffamily,
                outer sep=0,
                l sep=0,
                grow'=0,
                child anchor=west,
                parent anchor=south,
                anchor=west,
                calign=first,
                edge path={
                    \noexpand\path [draw, \forestoption{edge}]
                    (!u.south west) +(5pt,0) |- node[fill,inner sep=1.25pt] {} (.child anchor)\forestoption{edge label};
                },
                before typesetting nodes={
                    if n=1
                    {insert before={[,phantom]}}
                    {}
                },
                fit=band,
                before computing xy={l=10pt},
            }
            [If
                [Compare
                    [Subscript
                        [\tcbox{$\ell_0\colon \texttt{list}$}]
                        [Index
                            [\tcbox{$\ell_1\colon \texttt{int}$}]]]
                    [{\ttfamily Eq}]
                    [Num [\tcbox{$\ell_2$}]]]
                [stmt*
                    [Assign
                        [\tcbox{$\ell_3$}]
                        [BinOp
                            [\tcbox{$\ell_3$}]
                            [{\ttfamily Add}]
                            [Num [{\ttfamily 1}]]]]
                        [\tcbox{\dots}]]]
        \end{forest}
    \end{sideways}
    \caption{
        \emph{Top:} An overview of \system.
        A miner \circled{1} extracts common idioms from the programs in a given dataset.
        All the idiom occurrences in the dataset programs are \circled{2} marked as optional alternative grammar operators.
        The dataset with marked occurrences is used to \circled{3} train a neural generative model.
        At inference time, the model \circled{4} generates programs with named idioms, which are inlined before
        program execution.
        Note that idioms may have named subexpressions, may repeat, and may occur at any program level.
        For clarity, we typeset idioms using function-like syntax ${\idiom_j}(\ell_1, \dots,
        \ell_k)$ in this paper, although they are actually represented as AST fragments with no syntax. \\
        \emph{Bottom:} AST fragment representation of the idiom $\idiom_2$ in Python.
        Here \textsf{sans-serif} nodes are fixed non-terminals, \texttt{monospaced} nodes are fixed terminals,
        and \holed{boxed} nodes are named arguments.
    }
    \label{fig:intro:overview}
    \vspace{-0.1\baselineskip}
\end{figure*}

%% file: sections/overview.tex
\section{Background}%
\label{sec:overview}

\paragraph{Program Synthesis}%

We consider the following formulation of the \emph{program synthesis} problem.
Assume an underlying programming language $\dsl$ of programs.
Each program $P \in \dsl$ can be represented either as a sequence $\tok_1 \cdots \tok_{|P|}$ of its \emph{tokens}, or,
equivalently, as an \emph{abstract syntax tree (AST)}~$\ast$ parsed according to the context-free grammar (CFG)
$\grammar$ of the language $\dsl$.
The goal of a program synthesis model $\model\colon \spec \mapsto P$ is to generate a program~$P$ that
maximizes the conditional probability $ \Pr\left( P \mid \spec \right) $ \ie the most likely program given
the specification.
We also assume a training set $\dataset = \left\{ \langle \spec_j, P_j \rangle \right\}_{j=1}^{|\dataset|}$, sampled
from an unknown true distribution $\mathfrak{D}$, from which
we wish to estimate the conditional probability $ \Pr \left( P \mid \spec \right) $.

In this work, we consider general-purpose programming languages $\dsl$ with a known context-free grammar~$\grammar$
such as Python and SQL.
Each \emph{specification} $\spec$ is represented as
a \emph{natural language task description}, \ie a sequence of words $\nlseq = \word_1 \cdots \word_{|\nlseq|}$
(although the \system synthesizer can be conditioned on any other type of incomplete spec).
In principle, we do not impose any restrictions on the generative model~$\model$ apart from it being
able to emit syntactically valid programs.
However, as we detail in \Cref{sec:synthesis}, the \system framework is most easily implemented on top of
\emph{structural generative models} such as sequence-to-tree models~\citep{yin17acl} and graph neural
networks~\citep{li2015gated,exprgen}.

\paragraph{Code Idioms}%

Following \citet{allamanis2014mining}, we define code idioms as \emph{fragments}~$\idiom$ of valid ASTs~$\ast$ in the
CFG~$\grammar$, \ie trees of nonterminals and terminals from $\grammar$ that may occur as subtrees of valid parse trees
from $\grammar$.
The grammar $\grammar$ extended with a set of idiom fragments forms a \emph{tree substitution grammar} (TSG).
We also associate a non-unique \emph{label} $\lab$ with each nonterminal leaf in every idiom, and require
that every instantiation of an idiom $\idiom$ must have its identically-labeled nonterminals instantiated to identical
subtrees.
This enables the role of idioms as \emph{subroutines}, where labels act as ``named arguments'' in the ``body'' of an
idiom.
See \Cref{fig:intro:overview} for an example.

%% file: sections/mining.tex
\section{Mining Code Idioms}%
\label{sec:idioms}

The first step of \system is obtaining a set of frequent and useful AST fragments as code idioms.
The trade-off between frequency and usefulness is crucial: it is trivial to mine \emph{commonly occurring} short
patterns, but they are often meaningless~\citep{aggarwal2014frequent}.
Instead, we employ and extend the methodology of \citet{allamanis2018mining} and frame idiom mining as a nonparameteric
Bayesian problem.

We represent idiom mining as inference over \emph{probabilistic tree substitution grammars} (pTSG).
A pTSG is a probabilistic context-free grammar extended with production rules that expand to a whole AST fragment
instead of a single level of symbols~\citep{cohn2010inducing,post2009bayesian}.
The grammar $\grammar$ of our original language~$\dsl$ induces a pTSG~$\grammar_0$ with no fragment rules and with
choice probabilities estimated from the corpus~$\dataset$.
To construct a pTSG corresponding to the extension of $\dsl$ with common tree fragments representing idioms,
we define a distribution $\mathfrak{G}$ over pTSGs as follows.

We first choose a Pitman-Yor process~\citep{teh2010hierarchical} as a prior distribution $\mathfrak{G}_0$ over pTSGs.
It is a nonparameteric process that has proven to be effective for mining code idioms in prior work thanks to its
modeling of production choices as a Zipfian distribution (in other words, it implements the desired ``rich get richer''
effect, which encourages a smaller number of larger \emph{and} more common idioms).
\begin{floatingfigure}{0.45\textwidth}
    \vspace{-5pt}
    \includegraphics[width=0.45\textwidth, trim=80 18 18 18, clip]{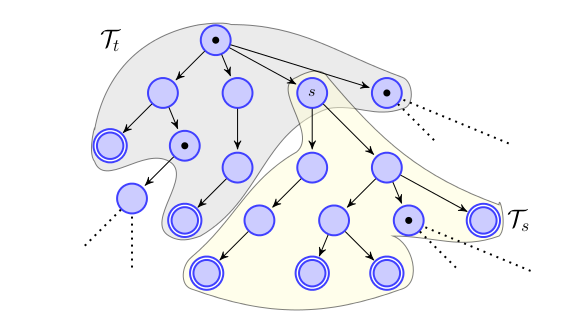}
    \caption{
        MCMC sampling for an AST (figure from~\cite{allamanis2014mining}).
        Dots show the inferred nodes where the AST is split into fragments.
    }
    \label{fig:mining:allamanis}
\end{floatingfigure}
Formally, it is a ``stick-breaking'' process~\citep{sethuraman1994constructive} that defines $\mathfrak{G}_0$ as a
distribution for each \emph{set of idioms~$\widetilde\idiom_N$ rooted at a nonterminal symbol $N$} as
\allowdisplaybreaks
\begin{align*}
    &\Pr (\idiom \in \widetilde\idiom_N) \bydef \sum_{k=0}^\infty \pi_k \, \delta\left(\idiom = \idiom_k\right), \quad
        \idiom_k \sim \grammar_0 \\
    &\pi_k \bydef u_k \prod_{j=1}^{k-1} (1 - u_j),\ u_k \sim \mathrm{Beta}\left( 1 - d,\, \alpha + k d \right)
\end{align*}
where $\delta(\cdot)$ is the delta function, and $\alpha$, $d$ are hyperparameters.
See \citet{allamanis2018mining} for details.

\system uses $\mathfrak{G}_0$ to compute a posterior distribution $\mathfrak{G}_1 = \Pr \left( \grammar_1 \mid \ast_1,
\dots, \ast_N \right)$ using Bayes' rule, where $\ast_1, \dots, \ast_N$ are concrete AST fragments in the training set
$\dataset$.
As this calculation is computationally intractable, we approximate it using type-based MCMC~\citep{liang2010type}.
At each iteration~$t$ of the MCMC process, \system generates a pTSG~$\grammar_t$ whose distribution approaches
$\mathfrak{G}_1$ as $t \to \infty$.
It works by sampling \emph{splitting points} for each AST $\ast$ in the corpus $\dataset$, which by construction define
a set of fragments constituting $\grammar_t$ (see \Cref{fig:mining:allamanis}).
The split probabilities of this Gibbs sampling are set in a way that incentivizes merging adjacent tree fragments that
often cooccur in $\dataset$.
The final idioms are then extracted from the pTSG obtained at the last MCMC iteration.

While the Pitman-Yor process helps avoid overfitting the idioms to $\dataset$, not all sampled idioms are useful for
synthesis.
Thus we \emph{rank} and \emph{filter} the idioms before using them in the training.
In this work, we reuse two ranking functions defined by \citet{allamanis2018mining}:
\begin{align*}
    \mathsf{Score_{Cov}} \left( \idiom \right) &\bydef
        \text{coverage} = \mathsf{count}( \ast \in \dataset \mid \idiom \in \ast ) \\
    \mathsf{Score_{CXE}} \left( \idiom \right) &\bydef
        \text{coverage} \cdot \text{cross-entropy gain}
        = \frac{ \mathsf{count}( \ast \in \dataset \mid \idiom \in \ast ) }{ |\dataset| } \cdot \frac{1}{|\idiom|} \log \frac{ \Pr_{\mathfrak{G}_1}(\idiom) }{ \Pr_{\mathfrak{G}_0}(\idiom) }
\end{align*}
and also filter out any \emph{terminal} idioms (\ie those that do not contain any named arguments $\lab$).

We conclude with a brief analysis of computational complexity of idiom mining.
Every iteration of the MCMC sampling traverses the entire dataset $\dataset$ once to sample the random variables that
define the splitting points in each AST.
When run for $M$ iterations, the complexity of idiom mining is $\order{M \cdot \sum_{\ast \in \dataset} |\ast|}$.
Idiom ranking adds an additional step with complexity $\order{|\widetilde\idiom| \log |\widetilde\idiom|}$ where
$\widetilde\idiom$ is the set of idioms obtained at the last iteration.
In our experiments (detailed in \Cref{sec:evaluation}) we set $M=10$, and the entire idiom mining takes less than 10
minutes on a dataset of $|\dataset| \approx \np{10000}$ ASTs.

%% file: sections/synthesis.tex
\section{Using Idioms in Program Synthesis}%
\label{sec:synthesis}

Given a set of common idioms $\widetilde\idiom = \{ \idiom_1, \dots, \idiom_N \}$ mined by $\system$,
we now aim to learn a synthesis model~$\model$ that emits whole idioms $\idiom_j$ as atomic actions instead of
individual AST nodes that comprise~$\idiom_j$.
Achieving this involves two key challenges.

First, since idioms are represented as AST fragments without concrete syntax, \system works best when the
synthesis model $\model$ is \emph{structural}, \ie it generates the program AST instead of its syntax.
Prior work~\cite{yin17acl,yin2018structvae,exprgen} also showed that tree- and graph-based code generation models
outperform sequence-to-sequence models, and thus we adopt a similar architecture in this work.

Second, exposing the model $\model$ to idiom usage patterns is not obvious.
One approach could be to extend the grammar with new named operators $\op_{\idiom}(\lab_1, \dots, \lab_k)$ for
each idiom $\idiom$, replace every occurrence of $\idiom$ with $\op_{\idiom}$ in the data, and
train the synthesizer on the rewritten dataset.
However, this would not allow $\model$ to learn from the idiom definitions (bodies).
In addition, idiom occurrences often overlap, and any deterministic rewriting strategy would arbitrarily discard some
occurrences from the corpus, thus limiting the model's exposure to idiom usage.
In our experiments, we found that greedy rewriting discarded as many as $75\%$ potential idiom occurrences from the
dataset.
Therefore, a successful training strategy must preserve all occurrences and instead let the model \emph{learn} a
rewriting strategy that optimizes end-to-end synthesis accuracy.

To this end, we present a novel training setup for code generation that encourages the model to choose the most
useful subset of idioms and the best representation of each program in terms of the idioms.
It works by
\begin{inlinelist}
    \item marking occurrences of the idioms $\widetilde\idiom$ in the training set $\dataset$
    \item at training time, encouraging the model to emit \emph{either} the whole idiom \emph{or} its
        body for every potential idiom occurrence in the AST
    \item at inference time, replacing the model's state after emitting an idiom~$\idiom$ with the state the model would
        have if it had emitted $\idiom$'s body step by step.
\end{inlinelist}

\subsection{Model Architecture}%
\label{sub:synthesis:model}

The synthesis model $\model$ of \system combines a \emph{spec encoder} $\encoder$ and an \emph{AST decoder}~$\decoder$,
following the formulation of \citet{yin17acl}.
The encoder $\encoder$ embeds the NL specification $\nlseq = \word_1 \cdots \word_n$ into
word representations $\nlseqemb = \wemb_1 \cdots \wemb_n$.
The decoder $\decoder$ uses an LSTM to model the sequential generation of the AST in the depth-first order,
wherein each timestep~$t$ corresponds to \emph{an action} $\action_t$ --- either (a) expanding a production
from the grammar, (b) expanding an idiom, or (c) generating a terminal token.
Thus, the probability of generating an AST $\ast$ given $\nlseqemb$ is
\begin{equation}
    \label{eq:synthesis:objprogram}
    \Pr(\ast \mid \nlseqemb) = \prod\nolimits_t \Pr\bigl(\action_t \mid \ast_{t}, \nlseqemb\bigr)
\end{equation}
where $\action_t$ is the action taken at timestep $t$, and $\ast_{t}$ is the partial AST generated before $t$.
The probability $\Pr(\action_t \mid \ast_{t}, \nlseqemb)$ is computed from the decoder's hidden state $\hidden_{t-1}$
depending on $\action_t$.

\paragraph{Production Actions}
For actions $\action_t = \textsc{ApplyRule}[\production]$ corresponding to expanding production rules $\production \in
\grammar$ from the original CFG $\grammar$, we compute the probability
$\Pr(\action_t \mid \ast_{t}, \nlseqemb)$ by encoding the current partial AST structure similarly to \citet{yin17acl}.
Specifically, we compute the new hidden state as
$ \hidden_t = f_{\text{LSTM}}\left( [\vect{\action}_{t-1} \conc \vect{c}_t \conc \hidden_{p_t} \conc
\vect{\action}_{p_t} \conc \vect{n}_{f_t}],\ \hidden_{t-1} \right) $
where $\vect{a}_{t-1}$ is the embedding of the previous action, $\vect{c}_t$ is the result of soft attention applied to
the spec embeddings~$\nlseqemb$ as per \citet{bahdanau2014neural}, $p_t$ is the timestep corresponding to expanding the
parent AST node of the current node, and $\vect{n}_{f_t}$ is the embedding of the current node type.
The hidden state $\hidden_t$ is then used to compute probabilities of the syntactically appropriate production rules
$\production \in \grammar$:
\begin{equation}
    \Pr(\action_t = \textsc{ApplyRule}[\production] \mid \ast_{t}, \nlseqemb) =
        \mathsf{softmax}_{\production}\left(g(\hidden_t)\right)
    \label{eq:synthesis:applyrule}
\end{equation}
where $g(\cdot)$ is a 2-layer MLP with a $\mathsf{tanh}$ non-linearity.

\paragraph{Terminal Actions}
For actions
$\action_t = \textsc{GetToken}[\tok]$, we compute the probability \mbox{$\Pr(\action_t \mid \ast_{t},\nlseqemb)$}
by combining a small vocabulary $\vocab$ of tokens commonly observed in the training data with a
\emph{copying mechanism}~\cite{hearthstone,see2017get} over the input $\nlseq$ to handle UNK tokens.
Specifically, we learn two
functions $p_\text{gen}(\hidden_t)$ and $p_\text{copy}(\hidden_t, \nlseq)$ such that
$p_\text{gen}$ produces a score for each vocabulary token $\tok \in \vocab$
and $p_{\text{copy}}$ computes a score for
copying the token $\tok$ from the input.
The scores are then normalized across the entries corresponding to the same constant, as in \cite{yin17acl,exprgen}.

\subsection{Training to Emit Idioms}%
\label{sec:synthesis:training}

As discussed earlier, training the model to emit idioms presents computational and learning challenges.
Ideally, we would like to extend \cref{eq:synthesis:objprogram} to maximize
\begin{equation}
    \label{eq:synthesis:objideal}
    \mathcal{J} =
        \sum_{\tau \in \mathcal{T}} \prod_{i=1}^{|\tau|} \Pr(\action_{\tau_i} \mid \ast_{\tau_i}, \nlseqemb)
\end{equation}
where $ \mathcal{T} $ is a set of different \emph{action traces} that may produce the output AST $\ast$.
The traces $\tau \in \mathcal{T}$ differ only in their possible choices of idiom actions
$\textsc{ApplyRule}[\op_\idiom]$ that emit some tree fragments of $\ast$ in a single step.
However, computing \cref{eq:synthesis:objideal} is intractable because idiom occurrences overlap and cause combinatorial
explosion in the number of traces~$\mathcal{T}$.
Instead, we apply Jensen's inequality and maximize a lower bound:
\begin{equation}
    \label{eq:synthesis:jensen}
        \log \mathcal{J} =
            \log \sum_{\tau \in \mathcal{T}} \prod_{i=1}^{|\tau|} \Pr(\action_{\tau_i} \mid \ast_{\tau_i}, \nlseqemb)
            \ge \log(|\mathcal{T}|) + \frac{1}{|\mathcal{T}|} \sum_{\tau \in \mathcal{T}} \sum_{i=1}^{|\tau|}
            \log \Pr(\action_{\tau_i} \mid \ast_{\tau_i}, \nlseqemb)
\end{equation}

Let $A(\ast_{t}) = \{ \action^*_t \} \cup I(\ast_{t})$ be the set of all valid actions to expand the
AST $\ast_{t}$ at timestep $t$.
Here $\action^*_t$ is the action from the \emph{original action trace} that generates $\ast$ using the original CFG and
$I(\ast_{t})$ is the set of idiom actions $\textsc{ApplyRule}[\op_\idiom]$ also applicable at the node to be expanded
in $\ast_{t}$.
Let $c(\mathcal{T}, t)$ also denote the number of traces $\tau \in \mathcal{T}$ that admit an action choice for the AST
$\ast_{t}$ from the original action trace.
Since each action $\action \in A(\ast_{t})$ occurs in the sum in \cref{eq:synthesis:jensen} with probability
${c(\mathcal{T}, t)} \,\big/\, {|A(\ast_{t})|}$, we can
rearrange this sum over traces as a sum over timesteps of the original trace:
\begin{align}
    \label{eq:synthesis:objactual}
    &\frac{1}{|\mathcal{T}|} \sum_{\tau \in \mathcal{T}} \sum_{i=1}^{|\tau|}
    \log \Pr(\action_{\tau_i} \mid \ast_{\tau_i}, \nlseqemb)
    = \frac{1}{|\mathcal{T}|} \sum_{t} \sum_{\action \in A(\ast_{t})}
    \frac{c(\mathcal{T}, t)}{|A(\ast_{t})|} \log \Pr(\action \mid \ast_{\tau_i}, \nlseqemb) \nonumber\\
    &\quad = \sum_{t} \frac{1}{|A(\ast_{t})|} \sum_{\action \in A(\ast_{t})}
    \frac{c(\mathcal{T}, t)}{|\mathcal{T}|} \log \Pr(\action \mid \ast_{\tau_i}, \nlseqemb)
    = \expect_{\ast_t \sim \mathcal{T}} \frac{1}{|A(\ast_{t})|} \sum_{\action \in A(\ast_{t})} \log \Pr(\action \mid
    \ast_{\tau_i}, \nlseqemb) \nonumber\\
    &\quad \approx \sum_{t} \frac{1}{|A(\ast_{t})|} \Bigl[ \log \Pr(\action^*_t \mid \ast_{t}, \nlseqemb) +
        \sum_{\mathclap{\idiom \in M(\ast_{t})}}
    \log \Pr(\action_t = \textsc{ApplyRule}[\op_\idiom] \mid \ast_{t}, \nlseqemb) \Bigr]
\end{align}
\begin{floatingfigure}{0.5\textwidth}
    \centering
    \tcbset{size=fbox}
    \vspace{-\baselineskip}
    \includegraphics[width=0.5\textwidth, trim=0 80pt 300pt 0, clip]{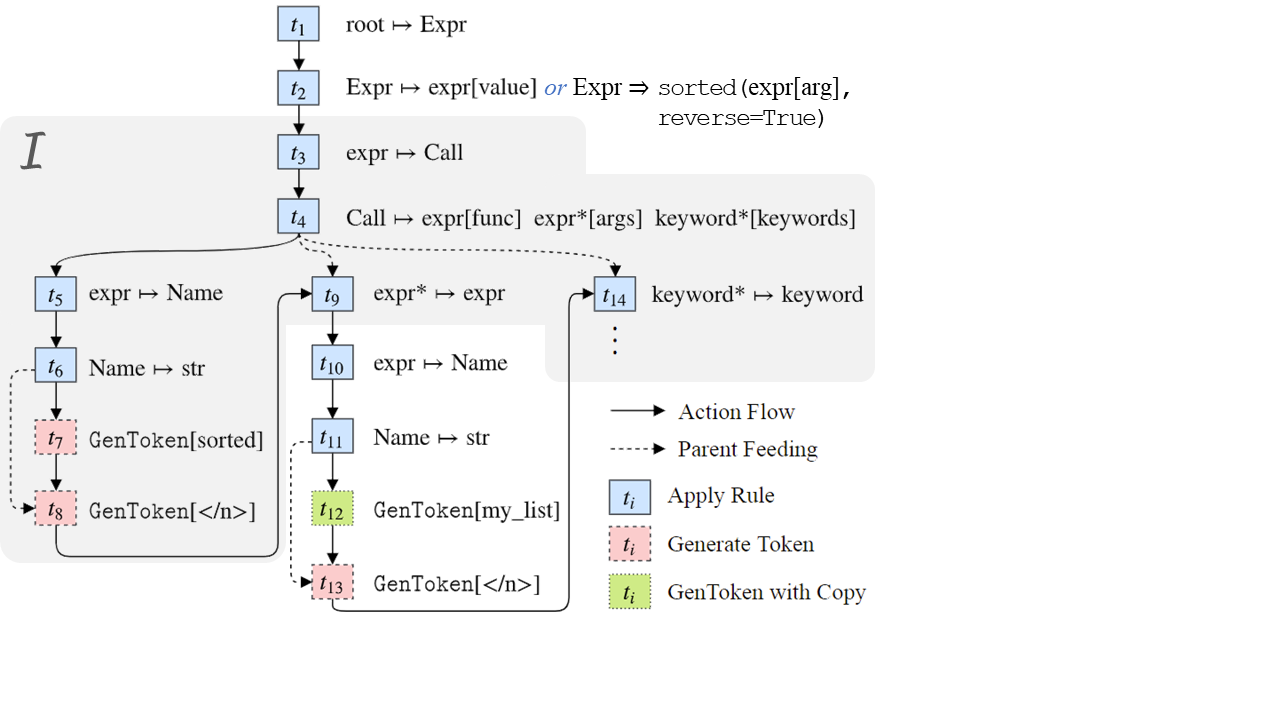}
    \caption{
        Decoding the AST \lstinline[language=Python]{sorted(my_list,reverse=True)}, figure adapted from
        \cite{yin17acl}.
        Suppose an idiom
        $\idiom=\ $\lstinline[language=Python]{sorted(}\holed{$\lab$},\,\lstinline[language=Python]{reverse=True)}
        is mined and added as an operator $\op_\idiom(\lab)$ to the grammar.
        At training time, \system adjusts the cross-entropy objective at timestep $t_2$ to additionally allow
        $\op_\idiom$ as a valid production, with no change to further decoding.
        At inference time, if decoder emits an action $a_{t_2} = \textsc{ApplyRule}[\op_\idiom]$, \system unrolls
        $\idiom$ on the fly by teacher-forcing the shaded portion of the AST generation.
    }
    \label{fig:synthesis:decoder}
\end{floatingfigure}
In the last step of \Cref{eq:synthesis:objactual}, we approximate the expectation over ASTs randomly drawn from all
traces~$\mathcal{T}$ using only the original trace (containing all possible $\ast_t$) as a Monte Carlo estimate.

Intuitively, at each timestep during training we encourage the model to emit \emph{either} the original AST action for
this timestep \emph{or} any applicable idiom that matches the AST at this step, with no penalty to either choice.
However, to avoid the combinatorial explosion, we only teacher-force the original generation trace (\emph{not} the idiom
bodies), thus optimizing the bound in \cref{eq:synthesis:objactual}.
\Cref{fig:synthesis:decoder} illustrates this optimization process on an example.

At inference time, whenever the model emits an $\textsc{ApplyRule}[\op_\idiom]$ action, we teacher-force the body of
$\idiom$ by substituting the embedding of the previous action $\vect{a}_{t-1}$ with embedding of the \emph{previous
action in the idiom definition}, thus emulating the tree fragment expansion.
Outside the bounds of $\idiom$ (\ie within the hole subtrees of $\idiom$) we use the actual $\vect{a}_{t-1}$ as usual.

%% file: sections/evaluation.tex
\section{Evaluation}%
\label{sec:evaluation}

\paragraph{Datasets}
We evaluate \system on two semantic parsing datasets:
Hearthstone~\citep{hearthstone}
and Spider~\citep{spider}.

Hearthstone is a dataset of 665 card descriptions from the trading
card game of the same name, along with the implementations of their effects in Python using the game APIs.
The descriptions act as NL specs $\nlseq$, and are on average 39.1 words long.

Spider is a dataset of \np{10181} questions describing \np{5693}
unique SQL queries over \np{200} databases with multiple tables each.
Each question pertains to a particular database, whose schema is given to the synthesizer.
Database schemas do not overlap between the train and test splits, thus challenging the model to generalize across
different domains.
The questions are on average 13 words long and databases have on average 27.6 columns and 8.8 foreign keys.

\paragraph{Implementation}

We mine the idioms using the training split of each dataset.
Thus \system cannot indirectly overfit to the test set by learning its idioms, but it also cannot generalize beyond the
idioms that occur in the training set.
We run type-based MCMC (\Cref{sec:idioms}) for \np{10} iterations with $\alpha = 5$ and $d = \np{0.5}$.
After ranking (with either $\mathsf{Score_{COV}}$ or $\mathsf{Score_{CXE}}$) and filtering, we use $K$
top-ranked idioms to train the generative model.
We ran ablation experiments with $K \in \left\{ 10, 20, 40, 80 \right\}$.

As described in \Cref{sec:synthesis}, for all our experiments we used a tree-based decoder with
a pointer mechanism as the synthesizer $\model$, which we implemented in PyTorch~\cite{pytorch}.
For the Hearthstone dataset, we use a bidirectional LSTM~\cite{lstm} to implement the description encoder $\nlseqemb =
\encoder(\nlseq)$, similarly to \citet{yin17acl}.
The word embeddings $\wemb$ and hidden LSTM states $\hidden$ have dimension \np{256}.
The models are trained using the Adadelta optimizer~\citep{zeiler2012adadelta} with learning rate $1.0$, $\rho = 0.95$,
$\varepsilon = \np{e-6}$ for up to \np{2600} steps with a batch size of 10.

For the Spider dataset,
word embeddings $\wemb$ have dimension \np{300}, and hidden LSTM states $\hidden$ have dimension \np{256}.
The models are trained using the Adam optimizer~\citep{adam} with $\beta_1 = 0.9$, $\beta_2 = 0.999$,
$\varepsilon=\np{e-9}$ for up to \np{40000} steps with a batch size of 10.
The learning rate warms up linearly up to \np{2.5e-4} during the first \np{2000} steps,
and then decays polynomially by $ \left( 1 - t / T \right)^{-0.5} $ where $T$ is the total number of steps.
Each model configuration is trained on one NVIDIA GTX 1080 Ti GPU.

The Spider tasks additionally include the \emph{database schema} as an input in the description.
We follow a recent approach of embedding the schema using relation-aware self-attention within the
encoder~\cite{spiderschema}.
Specifically, we initialize a representation for each column, table, and word in the question, and then update these
representations using 4 layers of relation-aware self-attention~\cite{shaw2018self} using a graph that describes the
relations between columns and tables in the schema.
See Section~\ref{sec:appendix-spider-encoder} in the appendix for more details about the Spider schema encoder.

\subsection{Experimental Results}

In each configuration, we compare the performance of equivalent trained models on the same dataset with and without
idiom-based training of \system.
For fairness, we show the performance of the same decoder implementation described in \Cref{sub:synthesis:model} as a
baseline rather than the state-of-the-art results achieved by different approaches from the literature.
Thus, our baseline is the decoder described in \Cref{sub:synthesis:model} trained with a regular cross-entropy objective
rather than the \system objective in \Cref{eq:synthesis:objactual}.
Following prior work, we evaluate program generation as a semantic parsing task, and measure
\textbf{(i)} exact match accuracy and BLEU scores for Hearthstone and
\textbf{(ii)} exact match accuracy of program sketches for Spider.

\newcommand{\cpbox}[1]{\pbox[t]{1.08cm}{\centering#1\vspace{2pt}}}

\begin{table}[t]
    \begin{minipage}[t]{0.56\textwidth}
        \centering
        \small
        \caption{Ablation tests on the Hearthstone dev set.\vspace{2pt}}
        \label{tab:results:hs}
        \begin{tabular}{llccc}
            \toprule
            \textbf{Model} & $\bm{K}$ &
                \cpbox{\textbf{Exact match}} &
                \cpbox{\textbf{Sentence BLEU}} &
                \cpbox{\textbf{Corpus BLEU}} \\
            \midrule
            Baseline decoder & --- & \textbf{0.197} & 0.767 & 0.763 \\
            \midrule
                \multirow{4}{*}{\system, $\mathsf{Score_{Cov}}$} & 10 & 0.151 & 0.781 & \textbf{0.785} \\
                                                                 & 20 & 0.091 & 0.745 & 0.745 \\
                                                                 & 40 & 0.167 & 0.765 & 0.764 \\
                                                                 & 80 & \textbf{0.197} & 0.780 & 0.774 \\
            \midrule
                \multirow{4}{*}{\system, $\mathsf{Score_{CXE}}$} & 10 & 0.151 & 0.780 & 0.783 \\
                                                                 & 20 & 0.167 & \textbf{0.787} & 0.782 \\
                                                                 & 40 & 0.182 & 0.773 & 0.770 \\
                                                                 & 80 & 0.151 & 0.771 & 0.768 \\
            \bottomrule
        \end{tabular}
    \end{minipage}
    \begin{minipage}[t]{0.48\textwidth}
        \centering
        \small
        \caption{Ablation tests on the Spider dev set.\vspace{2pt}}
        \label{tab:results:spider}
        \begin{tabular}{llr}
            \toprule
            \textbf{Model} & $\bm{K}$ & \textbf{Exact match} \\
            \midrule
            Baseline decoder & --- & 0.395 \\
            \midrule
            \multirow{4}{*}{\system, $\mathsf{Score_{Cov}}$} & 10 & 0.394 \\
                                                                 & 20 & 0.379 \\
                                                                 & 40 & 0.395 \\
                                                                 & 80 & 0.407 \\
                                                                 \midrule
            \multirow{4}{*}{\system, $\mathsf{Score_{CXE}}$} & 10 & 0.368 \\
                                                                 & 20 & 0.382 \\
                                                                 & 40 & 0.387 \\
                                                                 & 80 & \textbf{0.416} \\
                                                                 \bottomrule
        \end{tabular}
    \end{minipage}
\end{table}

\Cref{tab:results:hs,tab:results:spider} show our ablation analysis of different configurations of \system on the
Hearthstone and Spider dev sets, respectively.
\Cref{tab:results:test} shows the test set results of the best model configuration for Hearthstone
(the test instances for the Spider dataset are unreleased).
\begin{wraptable}[8]{r}{0.45\textwidth}
    \vspace{-0.6\baselineskip}
    \centering
    \small
    \caption{
        Test set results on Hearthstone (using the best configurations on the dev set).
    }
    \label{tab:results:test}
    \begin{tabular}{lccc}
        \toprule
        \textbf{Model} &
                \cpbox{\textbf{Exact match}} &
                \cpbox{\textbf{Sentence BLEU}} &
                \cpbox{\textbf{Corpus BLEU}} \\
        \midrule
        Baseline & 0.152 & 0.743 & 0.723 \\
        \system & \textbf{0.197} & \textbf{0.780} & \textbf{0.766} \\
        \bottomrule
    \end{tabular}
\end{wraptable}
As the results show, small numbers of idioms do not significantly change the exact match accuracy but improve BLEU
score, and $K=80$ gives a significant improvement in both the exact match accuracy and BLEU scores.
The improvement is even more pronounced on the test set with $4.5\%$ improvement in exact match accuracy and more
than 4 BLEU points, which shows that mined training set idioms generalize well to the whole data distribution.
As mentioned above, we compare only to the same baseline architecture for fairness, but \system could also be easily
implemented on top of the structural CNN decoder of \citet{sun2018grammar}, the current state of the art on the
Hearthstone dataset.

\begin{figure*}[t]
    \centering
    \tcbset{size=fbox}
    \lstset{gobble=9, escapeinside=<>, mathescape=true, basicstyle=\scriptsize\ttfamily,
            xleftmargin=-10pt, xrightmargin=-10pt, aboveskip=4pt, belowskip=4pt}
    \begin{minipage}[b]{0.34\linewidth}
        \begin{lstlisting}[language=Python]
           def __init__(self):
               super().__init__(<\holed{$\lab_0\colon \mathtt{str}$}>, <\holed{$\lab_1\colon \mathtt{int}$}>,
                   CHARACTER_CLASS.<\holed{$\lab_3\colon \mathtt{id}$}>,
                   CARD_RARITY.<\holed{$\lab_4\colon \mathtt{id}$}>, <\holed{$\lab_5^?$}>)
        \end{lstlisting}
    \end{minipage}
    \begin{minipage}[b]{0.20\linewidth}
        \begin{lstlisting}[language=Python]
             <\holed{$\lab_0\colon \mathtt{id}$}> =
               copy.copy(<\holed{$\lab_1\colon \mathtt{expr}$}>)
        \end{lstlisting}
        \hrule
        \begin{lstlisting}[language=Python]
            class <\holed{$\lab_0\colon \mathtt{id}$}>(<\holed{$\lab_1\colon \mathtt{id}$}>):
                def __init__(self):
        \end{lstlisting}
    \end{minipage}
    \hspace{0.02\linewidth}
    \begin{minipage}[b]{0.34\linewidth}
        \begin{lstlisting}[language=SQL]
            SELECT COUNT(<\holed{$\lab_0\colon \mathtt{col}$}>), <\holed{$\lab_1^*$}> WHERE <\holed{$\lab_2^*$}>
            INTERSECT <\holed{$\lab_4^?\colon \mathtt{sql}$}> EXCEPT <\holed{$\lab_5^?\colon \mathtt{sql}$}>
        \end{lstlisting}
        \hrule
        \begin{lstlisting}[language=SQL]
            WHERE <\holed{$\lab_0\colon \mathtt{col}$}> = <\$>terminal
        \end{lstlisting}
    \end{minipage}
    \caption{Five examples of commonly used idioms from the Hearthstone and Spider datasets.}
    \label{fig:evaluation:examples}
\end{figure*}

\Cref{fig:evaluation:examples} shows some examples of idioms that were frequently used by the model.
On Hearthstone, the most popular idioms involve common syntactic elements (\eg class and function definitions) and
domain-specific APIs commonly used in card implementations (\eg \texttt{CARD\_RARITY} enumerations or
\texttt{copy.copy} calls).
On Spider, they capture the most common combinations of SQL syntax, such as a \lstinline[language=SQL]{SELECT} query
with a single \lstinline[language=SQL]{COUNT} column and optional \lstinline[language=SQL]{INTERSECT} or
\lstinline[language=SQL]{EXCEPT} clauses.
Notably, popular idioms are also often \emph{big}: for instance, the first idiom in \Cref{fig:evaluation:examples}
expands to a tree fragment with more than 20 nodes.
Emitting it in a single step vastly simplifies the decoding process.

\begin{figure}[t]
    \centering
    \includegraphics[scale=0.42]{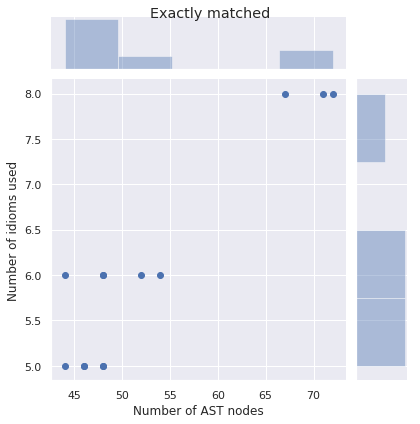}
    ~
    \includegraphics[scale=0.42]{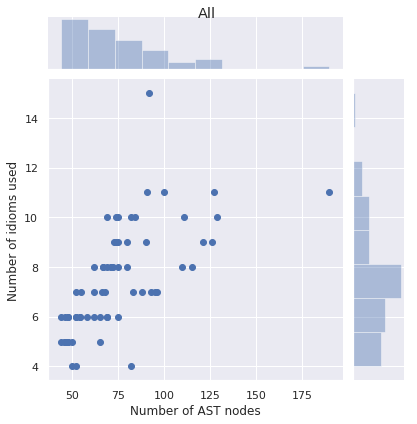}
    \caption{
        The distribution of used idioms in the inferred ASTs on the Hearthstone test set.
        \emph{Left:} in the ASTs exactly matched with ground truth; \emph{Right:} all ASTs.
    }%
    \label{fig:evaluation:usage}
    \vspace{-\baselineskip}
\end{figure}

We further conducted qualitative experiments to analyze actual idiom usage by \system on the Hearthstone test set.
\Cref{fig:evaluation:usage} shows the distribution of idioms used in the inferred (not ground truth) ASTs.
A typical program involves 7 idioms on average, or 6 for the programs that exactly match the ground truth.
Despite the widespread usage of idioms, not all of the mined idioms $\widetilde\idiom$ were useful: only 51 out of
$K=80$ idioms appear in the inferred ASTs.
This highlights the need for an end-to-end version of \system where idiom mining would be directly optimized to benefit
synthesis.

%% file: sections/related.tex
\section{Related Work}%
\label{sec:related}

\paragraph{Program synthesis \& Semantic parsing}

Program synthesis from natural language and input-output examples has a long history in Programming Languages (PL) and
Machine Learning (ML) communities (see \citet{gulwani2017survey} for a survey).
When an input specification is limited to natural language, the resulting problem can be considered \emph{semantic
parsing}~\citep{liang2016learning}.
There has been a lot of recent interest in applying recurrent sequence-based and tree-based neural networks to semantic
parsing~\citep{yin17acl,li2015gated,dong2016language,jia2016data,yin2018structvae}.
These approaches commonly use insights from the PL literature, such as grammar-based constraints to reduce the search
space, non-deterministic training oracles to enable multiple executable interpretations of intent, and supervision from
program execution.
They typically either supervise the training on one or more golden programs, or use reinforcement learning to supervise
the training from a neural program execution result~\citep{neelakantan2016learning}.
Our \system approach is applicable to any underlying neural semantic parsing model, as long as it is supervised by a
corpus of golden programs.
It is, however, most easily applicable to tree-based and graph-based models, which directly emit the AST of the target
program.
In this work we have evaluated \system as applied on top of the sequence-to-tree decoder of \citet{yin17acl}, and
extended it with a novel training regime that teaches the decoder to emit idiom operators in place of the idiomatic
code fragments.

\paragraph{Sketch generation}
Two recent works~\citep{coarse2fine,bayou} learn abstractions of the target program to compress and abstract the
reasoning process of a neural synthesizer.
Both of them split the generation process into \emph{sketch generation} and \emph{sketch completion}, wherein the first
stage emits a partial tree/sequence (\ie a \emph{sketch} of the program) and the second stage fills in the holes in this
sketch.
While sketch generation is typically implemented with a neural model, sketch completion can be either a different neural
model or a combinatorial search.
In contrast to \system, both works define the grammar of sketches manually by a deterministic \emph{program abstraction}
procedure and only allow a single top-level sketch for each program.
In addition, an earlier work of \citet{bovsnjak2017programming} also formulates program synthesis as sketch completion,
but in their work program sketches are manually provided rather than learned.
In \system, we learn the abstractions (code idioms) automatically from a corpus and allow them to appear
anywhere in the program, as is common in real-life programming.

\paragraph{Learning abstractions}

Recently, \citet{scc} developed an Explore, Compress \& Compile~(EC$^2$) framework for automatically
learning DSLs for program synthesis from I/O examples (such as the DSLs used by FlashFill~\citep{flashfill} and
DeepCoder~\citep{deepcoder}).
The workflow of EC$^2$ is similar to \system, with three stages:
\textbf{(a)} learn new DSL subroutines from a corpus of tasks,
\textbf{(b)} train a recognition model that maps a task specification to a distribution over DSL operators
as in DeepCoder~\citep{deepcoder}, and
\textbf{(c)} use these operators in a program synthesizer.
\system differs from EC$^2$ in three aspects:
\textbf{(i)} we assume a natural language specification instead of examples,
\textbf{(ii)} to handle NL specifications, our synthesizer is a neural semantic parser instead of enumerative
search, and
\textbf{(iii)} most importantly, we discover idioms that compress general-purpose languages instead of
extending DSLs.
Unlike for inductive synthesis DSLs such as FlashFill, the existence of \emph{useful} DSL abstractions for
general-purpose languages is not obvious, and our work is the first to demonstrate them.

Concurrently with this work, \citet{iyer2019learning} developed a different approach of learning code idioms for
semantic parsing.
They mine the idioms using a variation of \emph{byte-pair encoding} (BPE) compression extended to ASTs and greedily
rewrite all the dataset ASTs in terms of the found idioms for training.
While the BPE-based idiom mining is more computationally efficient than non-parametric Bayesian inference of \system,
introducing ASTs greedily tends to lose information about overlapping idioms, which we address in \system using our
novel training objective described in \Cref{sec:synthesis:training}.

As described previously, our code idiom mining is an extension of the procedure developed by
\citeauthor{allamanis2018mining}~\cite{allamanis2014mining,allamanis2018mining}.
They are the first to use the tree substitution grammar formalism and Bayesian inference to find non-trivial common
idioms in a corpus of code.
However, their problem formalization does not involve any application for the learned idioms beyond their explanatory
power.

%% file: sections/conclusion.tex
\section{Conclusion}%
\label{sec:conclusion}

Semantic parsing, or neural program synthesis from natural language, has made tremendous progress over the past years,
but state-of-the-art models still struggle with program generation at multiple levels of abstraction.
In this work, we present a framework that allows incorporating learned coding patterns from a corpus into the vocabulary
of a neural synthesizer, thus enabling it to emit high-level or low-level program constructs
interchangeably at each generation step.
Our current instantiation, \system, uses Bayesian inference to mine common code idioms, and employs a
novel nondeterministic training regime to teach a tree-based generative model to optionally emit whole idiom
fragments.
Such dataset abstraction using idioms improves the performance of neural program synthesis.

\system is only the first step toward learned abstractions in program synthesis.
While code idioms often correlate with latent semantic concepts
and our training regime allows the model to learn which idioms to use and in which context,
our current method does not mine them with the intent to directly optimize their usefulness for generation.
In future work, we want to alleviate this by jointly learning the mining and synthesis models, thus
optimizing the idioms' usefulness for synthesis by construction.
We also want to incorporate \emph{program semantics} into the idiom definition, such as data flow patterns or
natural language phrases from task specs.

%% file: sections/appendix-spider-encoder.tex
\section{Encoder for Spider dataset}
\label{sec:appendix-spider-encoder}

\begin{table}[t]
\caption{Description of edge types present in the directed graph created to represent the schema. An edge exists from node $x$ to node $y$ if the pair fulfills one of the descriptions listed in the table, with the corresponding label. Otherwise, no edge exists from $x$ to $y$.}
\label{table:schema-graph-edges}
\centering
\begin{tabular}{lllp{7cm}}
\toprule
Type of $x$ & Type of $y$ & Edge label & Description \\
\midrule
\multirow{3}{*}{Column} & \multirow{3}{*}{Column}
   & \textsc{Same-Table}    & $x$ and $y$ belong to the same table. \\
 & & \textsc{Foreign-Key-Col-F} & $x$ is a foreign key for $y$. \\
 & & \textsc{Foreign-Key-Col-R} & $y$ is a foreign key for $x$. \\
\midrule
\multirow{2}{*}{Column} & \multirow{2}{*}{Table}
   & \textsc{Primary-Key-F}   & $x$ is the primary key of $y$. \\
 & & \textsc{Belongs-To-F}    & $x$ is a column of $y$ (but not the primary key). \\
\midrule
\multirow{2}{*}{Table} & \multirow{2}{*}{Column}
   & \textsc{Primary-Key-R}   & $y$ is the primary key of $x$. \\
 & & \textsc{Belongs-To-R}    & $y$ is a column of $x$ (but not the primary key). \\
\midrule
\multirow{3}{*}{Table} & \multirow{3}{*}{Table}
   & \textsc{Foreign-Key-Tab-F}   & Table $x$ has a foreign key column in $y$. \\
 & & \textsc{Foreign-Key-Tab-R}   & Same as above, but $x$ and $y$ are reversed. \\
 & & \textsc{Foreign-Key-Tab-B}   & $x$ and $y$ have foreign keys in both directions. \\
\bottomrule
\end{tabular}
\end{table}

In the Spider dataset, each entry contains a question along with a database schema, containing tables and columns.
We will use the following notation:
\begin{itemize}
    \item $c_i$ for each column in the schema. Each column contains words $c_{i,1}, \cdots, c_{i,|c_i|}$.
    \item $t_i$ for each table in the schema. Each table contains words $t_{i,1}, \cdots, t_{i,|t_i|}$.
    \item $q$ for the input question. The question contains words $q_1, \cdots, q_{|q|}$.
\end{itemize}

\subsection{Encoding the Schema as a Graph}
\label{sec:encoding-as-graph}
We begin by representing the database schema using a directed graph $\mathcal{G}$, where each node and edge has a label.
We represent each table and column in the schema as a node in this graph, labeled with the words in the name;
for columns, we prepend the type of the column to the label.
For each pair of nodes $x$ and $y$ in the graph, Table~\ref{table:schema-graph-edges} describes when there exists an edge from $x$ to $y$ and the label it should have.

\subsection{Initial Encoding of the Input}
\label{sec:initial-encoding}
We now obtain an initial representation for each of the nodes in the graph, as well as for the words in the input question.
Formally, we perform the following:
\begin{align*}
(\mathbf{c}_{i,0}^\text{fwd}, \mathbf{c}_{i,0}^\text{rev}), \cdots, (\mathbf{c}_{i,|c_i|}^\text{fwd}, \mathbf{c}_{i,|c_i|}^\text{rev})
& = \text{BiLSTM}_\text{Column}(c_{i}^\text{type}, c_{i,1}, \cdots, c_{i,|c_i|}); \quad
\mathbf{c}_{i}^\text{init} = \text{Concat}(\mathbf{c}_{i,|c_i|}^\text{fwd}, \mathbf{c}_{i,0}^\text{rev}) \\
(\mathbf{t}_{i,1}^\text{fwd}, \mathbf{t}_{i,1}^\text{rev}), \cdots, (\mathbf{t}_{i,|t_i|}^\text{fwd}, \mathbf{t}_{i,|t_i|}^\text{rev})
& = \text{BiLSTM}_\text{Table}(t_{i,1}, \cdots, t_{i,|t_i|}); \quad
\mathbf{t}_{i}^\text{init} = \text{Concat}(\mathbf{t}_{i,|c_i|}^\text{fwd}, \mathbf{t}_{i,1}^\text{rev}) \\
(\mathbf{q}_{1}^\text{fwd}, \mathbf{q}_{1}^\text{rev}), \cdots, (\mathbf{q}_{|q|}^\text{fwd}, \mathbf{q}_{|q|}^\text{rev})
& = \text{BiLSTM}_\text{Question}(q_{1}, \cdots, q_{|q|}); \quad
\mathbf{q}_{i}^\text{init} = \text{Concat}(\mathbf{q}_{i}^\text{fwd}, \mathbf{q}_{i}^\text{rev})
\end{align*}
where each of the BiLSTM functions first lookup word embeddings for each of the input tokens.
The LSTMs do not share any parameters with each other.

\subsection{Relation-Aware Self-Attention}
\label{sec:rel-attn}
At this point, we have representations $\mathbf{c}_{i}^\text{init}$, $\mathbf{t}_{i}^\text{init}$, and $\mathbf{q}_{i}^\text{init}$.
Now, we would like to imbue these representations with the information in the schema graph.
We use a form of self-attention \citep{vaswaniAttentionAllYou2017} that is relation-aware \citep{shawSelfAttentionRelativePosition2018} to achieve this goal.

In one step of relation-aware self-attention, we begin with an input $\mathbf{x}$ of $n$ elements (where $x_i \in \mathbb{R}^{d_x}$) and transform each $x_i$ into $y_i \in \mathbb{R}^{d_z}$.
We follow the formulation described in \citet{shawSelfAttentionRelativePosition2018}:
\begin{align*}
    e_{ij}^{(h)} &= \frac{x_i W_Q^{(h)} (x_j W_K^{(h)} + \mathbf{\color{red}r_{ij}^K})^T}{\sqrt{d_z / H}}; \quad
    \alpha_{ij}^{(h)} = \frac{\exp(e_{ij}^{(h)})}{\sum_{l=1}^n \exp(e_{il}^{(h)})} \\
    z_i^{(h)} &= \sum_{j=1}^n \alpha_{ij}^{(h)} (x_j W_V^{(h)} + \mathbf{\color{red}r_{ij}^V}); \quad
    z_i = \text{Concat}(z_i^{(0)}, \cdots, z_i^{(H)}) \\
    \tilde{y}_i &= \text{LayerNorm}(x_i + z_i); \quad
    y_i = \text{LayerNorm}(\tilde{y}_i + \text{FC}(\text{ReLU}(\text{FC}(\tilde{y}_i)))
\end{align*}
The $r_{ij}$ terms encode the relationship between the two elements $x_i$ and $x_j$ in the input.
We explain how we obtain $r_{ij}$ in the next part.

For the application within the Spider encoder, we first construct the input $x$ of $|c| + |t| + |q|$ elements using $\mathbf{c}_{i}^\text{init}$, $\mathbf{t}_{i}^\text{init}$, and $\mathbf{q}_{i}^\text{init}$:
\[
    x = (\mathbf{c}_{1}^\text{init}, \cdots, \mathbf{c}_{|c|}^\text{init},
         \mathbf{t}_{1}^\text{init}, \cdots, \mathbf{t}_{|t|}^\text{init}, 
         \mathbf{q}_{1}^\text{init}, \cdots, \mathbf{q}_{|q|}^\text{init}).
\]
We then apply a stack of 4 relation-aware self-attention layers.
We set $d_z = d_x$ to facilitate this stacking.
The weights of the encoder layers are not tied; each layer has its own set of weights.

We define a discrete set of possible relation types, and map each type to an embedding to obtain $r_{ij}^V$ and $r_{ij}^K$.
We need a value of $r_{ij}$ for every pair of elements in $x$.
If $x_i$ and $x_j$ both correspond to nodes in $\mathcal{G}$ (i.e. each is either a column or table) with an edge from $x_i$ to $x_j$, then we use the label on that edge (possibilities listed in Table~\ref{table:schema-graph-edges}).

However, this is not sufficient to obtain $r_{ij}$ for every pair of $i$ and $j$.
In the graph we created for the schema, we have no nodes corresponding to the question words;
not every pair of nodes in the graph has an edge between them (the graph is not complete);
and we have no self-edges (for when $i = j$).
As such, we add more types beyond what is defined in Table~\ref{table:schema-graph-edges}:
\begin{itemize}
    \item $x_i \in $ question, $x_j \in $ question: 
       \textsc{Question-Dist-$d$}, where $d = \text{clip}(j - i, D)$; $\text{clip}(a, D) = \max(-D, \min(D, a))$. We use $D = 2$.
    \item If $i = j$, then \textsc{Column-Identity} or \textsc{Table-Identity}.
    \item $x_i \in $ question, $x_j \in \text{column} \cup \text{table}$; or $x_i \in \text{column} \cup \text{table}$, $x_j \in $ question: \\
       \textsc{Question-Column}, \textsc{Question-Table}, \textsc{Column-Question} or \textsc{Table-Question}
        depending on the type of $x_i$ and $x_j$.
    \item Otherwise, one of \textsc{Column-Column}, \textsc{Column-Table}, \textsc{Table-Column}, or \textsc{Table-Table}.
\end{itemize}
In the end, we add $2 + 5 + 4 + 4$ types beyond the $10$ in Table~\ref{table:schema-graph-edges}, for a total of 25 types.

After processing through the stack of $N$ encoder layers, we obtain
\begin{equation*}
     (\mathbf{c}_{1}^\text{final}, \cdots, \mathbf{c}_{|c|}^\text{final},
      \mathbf{t}_{1}^\text{final}, \cdots, \mathbf{t}_{|t|}^\text{final}, 
      \mathbf{q}_{1}^\text{final}, \cdots, \mathbf{q}_{|q|}^\text{final}) 
     = y.
\end{equation*}
We use $\mathbf{c}_{i}^\text{final}$, $\mathbf{t}_{i}^\text{final}$, and $\mathbf{q}_{i}^\text{final}$
in the decoder.

%% file: learned-syn-idioms.bbl
\begin{thebibliography}{42}
\providecommand{\natexlab}[1]{#1}
\providecommand{\url}[1]{\texttt{#1}}
\expandafter\ifx\csname urlstyle\endcsname\relax
  \providecommand{\doi}[1]{doi: #1}\else
  \providecommand{\doi}{doi: \begingroup \urlstyle{rm}\Url}\fi

\bibitem[Aggarwal and Han(2014)]{aggarwal2014frequent}
C.~C. Aggarwal and J.~Han.
\newblock \emph{Frequent pattern mining}.
\newblock Springer, 2014.

\bibitem[Allamanis and Sutton(2014)]{allamanis2014mining}
M.~Allamanis and C.~Sutton.
\newblock Mining idioms from source code.
\newblock In \emph{Proceedings of the \nth{22} {ACM} {SIGSOFT} International
  Symposium on Foundations of Software Engineering ({FSE})}, pages 472--483.
  ACM, 2014.

\bibitem[Allamanis et~al.(2018)Allamanis, Barr, Bird, Devanbu, Marron, and
  Sutton]{allamanis2018mining}
M.~Allamanis, E.~T. Barr, C.~Bird, P.~Devanbu, M.~Marron, and C.~Sutton.
\newblock Mining semantic loop idioms.
\newblock \emph{IEEE Transactions on Software Engineering}, 2018.

\bibitem[Bahdanau et~al.(2015)Bahdanau, Cho, and Bengio]{bahdanau2014neural}
D.~Bahdanau, K.~Cho, and Y.~Bengio.
\newblock Neural machine translation by jointly learning to align and
  translate.
\newblock In \emph{Proceedings of the \nth{3} International Conference on
  Learning Representations ({ICLR})}, 2015.

\bibitem[Balog et~al.(2017)Balog, Gaunt, Brockschmidt, Nowozin, and
  Tarlow]{deepcoder}
M.~Balog, A.~L. Gaunt, M.~Brockschmidt, S.~Nowozin, and D.~Tarlow.
\newblock {DeepCoder}: Learning to write programs.
\newblock In \emph{Proceedings of the \nth{5} International Conference on
  Learning Representations ({ICLR})}, 2017.

\bibitem[Bo{\v{s}}njak et~al.(2017)Bo{\v{s}}njak, Rockt{\"a}schel, Naradowsky,
  and Riedel]{bovsnjak2017programming}
M.~Bo{\v{s}}njak, T.~Rockt{\"a}schel, J.~Naradowsky, and S.~Riedel.
\newblock Programming with a differentiable {Forth} interpreter.
\newblock In \emph{Proceedings of the \nth{34} International Conference on
  Machine Learning ({ICML})}, volume~70, pages 547--556, 2017.

\bibitem[Brockschmidt et~al.(2019)Brockschmidt, Allamanis, Gaunt, and
  Polozov]{exprgen}
M.~Brockschmidt, M.~Allamanis, A.~L. Gaunt, and O.~Polozov.
\newblock Generative code modeling with graphs.
\newblock In \emph{Proceedings of the \nth{7} International Conference on
  Learning Representations ({ICLR})}, 2019.

\bibitem[Cohn et~al.(2010)Cohn, Blunsom, and Goldwater]{cohn2010inducing}
T.~Cohn, P.~Blunsom, and S.~Goldwater.
\newblock Inducing tree-substitution grammars.
\newblock \emph{Journal of Machine Learning Research}, 11\penalty0
  (Nov):\penalty0 3053--3096, 2010.

\bibitem[Devlin et~al.(2017{\natexlab{a}})Devlin, Bunel, Singh, Hausknecht, and
  Kohli]{karel}
J.~Devlin, R.~Bunel, R.~Singh, M.~Hausknecht, and P.~Kohli.
\newblock Neural program meta-induction.
\newblock In \emph{Advances in Neural Information Processing Systems ({NIPS})},
  pages 2080--2088, 2017{\natexlab{a}}.

\bibitem[Devlin et~al.(2017{\natexlab{b}})Devlin, Uesato, Bhupatiraju, Singh,
  Mohamed, and Kohli]{robustfill}
J.~Devlin, J.~Uesato, S.~Bhupatiraju, R.~Singh, A.-r. Mohamed, and P.~Kohli.
\newblock {RobustFill}: Neural program learning under noisy {I/O}.
\newblock In \emph{Proceedings of the \nth{34} International Conference on
  Machine Learning ({ICML})}, 2017{\natexlab{b}}.

\bibitem[Dong and Lapata(2016)]{dong2016language}
L.~Dong and M.~Lapata.
\newblock Language to logical form with neural attention.
\newblock In \emph{Proceedings of the \nth{54} Annual Meeting of the
  Association for Computational Linguistics ({ACL})}, 2016.

\bibitem[Dong and Lapata(2018)]{coarse2fine}
L.~Dong and M.~Lapata.
\newblock Coarse-to-fine decoding for neural semantic parsing.
\newblock In \emph{Proceedings of the \nth{56} Annual Meeting of the
  Association for Computational Linguistics ({ACL})}, 2018.

\bibitem[Ellis et~al.(2018)Ellis, Morales, Sabl\'{e}-Meyer, Solar-Lezama, and
  Tenenbaum]{scc}
K.~Ellis, L.~Morales, M.~Sabl\'{e}-Meyer, A.~Solar-Lezama, and J.~Tenenbaum.
\newblock Learning libraries of subroutines for neurally-guided {Bayesian}
  program induction.
\newblock In \emph{Advances in Neural Information Processing Systems}, pages
  7816--7826, 2018.

\bibitem[Gulwani(2011)]{flashfill}
S.~Gulwani.
\newblock Automating string processing in spreadsheets using input-output
  examples.
\newblock In \emph{Proceedings of the \nth{38} ACM Symposium on Principles of
  Programming Languages ({POPL})}, volume~46, pages 317--330, 2011.

\bibitem[Gulwani et~al.(2017)Gulwani, Polozov, and Singh]{gulwani2017survey}
S.~Gulwani, O.~Polozov, and R.~Singh.
\newblock Program synthesis.
\newblock \emph{Foundations and Trends\textsuperscript{\textregistered} in
  Programming Languages}, 4\penalty0 (1-2):\penalty0 1--119, 2017.

\bibitem[Hochreiter and Schmidhuber(1997)]{lstm}
S.~Hochreiter and J.~Schmidhuber.
\newblock Long short-term memory.
\newblock \emph{Neural computation}, 9\penalty0 (8):\penalty0 1735--1780, 1997.

\bibitem[Iyer et~al.(2019)Iyer, Cheung, and Zettlemoyer]{iyer2019learning}
S.~Iyer, A.~Cheung, and L.~Zettlemoyer.
\newblock Learning programmatic idioms for scalable semantic parsing.
\newblock In \emph{{EMNLP}}, 2019.

\bibitem[Jia and Liang(2016)]{jia2016data}
R.~Jia and P.~Liang.
\newblock Data recombination for neural semantic parsing.
\newblock In \emph{Proceedings of the \nth{54} Annual Meeting of the
  Association for Computational Linguistics ({ACL})}, volume~1, pages 12--22,
  2016.

\bibitem[Kalyan et~al.(2018)Kalyan, Mohta, Polozov, Batra, Jain, and
  Gulwani]{ngds}
A.~Kalyan, A.~Mohta, O.~Polozov, D.~Batra, P.~Jain, and S.~Gulwani.
\newblock Neural-guided deductive search for real-time program synthesis from
  examples.
\newblock In \emph{Proceedings of the \nth{6} International Conference on
  Learning Representations ({ICLR})}, 2018.

\bibitem[Kingma and Ba(2015)]{adam}
D.~P. Kingma and J.~Ba.
\newblock Adam: A method for stochastic optimization.
\newblock In \emph{Proceedings of \nth{3} International Conference on Learning
  Representations ({ICLR})}, 2015.

\bibitem[Li et~al.(2016)Li, Tarlow, Brockschmidt, and Zemel]{li2015gated}
Y.~Li, D.~Tarlow, M.~Brockschmidt, and R.~Zemel.
\newblock Gated graph sequence neural networks.
\newblock In \emph{Proceedings of the \nth{4} International Conference on
  Learning Representations ({ICLR})}, 2016.

\bibitem[Liang(2016)]{liang2016learning}
P.~Liang.
\newblock Learning executable semantic parsers for natural language
  understanding.
\newblock \emph{Communications of the ACM}, 59\penalty0 (9):\penalty0 68--76,
  2016.

\bibitem[Liang et~al.(2010)Liang, Jordan, and Klein]{liang2010type}
P.~Liang, M.~I. Jordan, and D.~Klein.
\newblock Type-based {MCMC}.
\newblock In \emph{Human Language Technologies: The 2010 Annual Conference of
  the North American Chapter of the Association for Computational Linguistics},
  pages 573--581. Association for Computational Linguistics, 2010.

\bibitem[Ling et~al.(2016)Ling, Blunsom, Grefenstette, Hermann,
  Ko{\v{c}}isk{\`y}, Wang, and Senior]{hearthstone}
W.~Ling, P.~Blunsom, E.~Grefenstette, K.~M. Hermann, T.~Ko{\v{c}}isk{\`y},
  F.~Wang, and A.~Senior.
\newblock Latent predictor networks for code generation.
\newblock In \emph{{ACL}}, volume~1, pages 599--609, 2016.
\newblock URL \url{https://github.com/deepmind/card2code}.

\bibitem[Murali et~al.(2018)Murali, Qi, Chaudhuri, and Jermaine]{bayou}
V.~Murali, L.~Qi, S.~Chaudhuri, and C.~Jermaine.
\newblock Neural sketch learning for conditional program generation.
\newblock In \emph{Proceedings of the \nth{6} International Conference on
  Learning Representations ({ICLR})}, 2018.

\bibitem[Neelakantan et~al.(2017)Neelakantan, Le, Abadi, McCallum, and
  Amodei]{neelakantan2016learning}
A.~Neelakantan, Q.~V. Le, M.~Abadi, A.~McCallum, and D.~Amodei.
\newblock Learning a natural language interface with neural programmer.
\newblock In \emph{Proceedings of the \nth{5} International Conference on
  Learning Representations ({ICLR})}, 2017.

\bibitem[Paszke et~al.(2017)Paszke, Gross, Chintala, Chanan, Yang, DeVito, Lin,
  Desmaison, Antiga, and Lerer]{pytorch}
A.~Paszke, S.~Gross, S.~Chintala, G.~Chanan, E.~Yang, Z.~DeVito, Z.~Lin,
  A.~Desmaison, L.~Antiga, and A.~Lerer.
\newblock Automatic differentiation in {PyTorch}.
\newblock 2017.

\bibitem[Polozov and Gulwani(2015)]{flashmeta}
O.~Polozov and S.~Gulwani.
\newblock {FlashMeta}: A framework for inductive program synthesis.
\newblock In \emph{Proceedings of the 2015 {ACM} {SIGPLAN} International
  Conference on Object-Oriented Programming, Systems, Languages, and
  Applications ({OOPSLA})}, pages 107--126, 2015.

\bibitem[Post and Gildea(2009)]{post2009bayesian}
M.~Post and D.~Gildea.
\newblock Bayesian learning of a tree substitution grammar.
\newblock In \emph{Proceedings of the ACL-IJCNLP 2009 Conference Short Papers},
  pages 45--48. Association for Computational Linguistics, 2009.

\bibitem[See et~al.(2017)See, Liu, and Manning]{see2017get}
A.~See, P.~J. Liu, and C.~D. Manning.
\newblock Get to the point: Summarization with pointer-generator networks.
\newblock In \emph{Proceedings of the \nth{55} Annual Meeting of the
  Association for Computational Linguistics ({ACL})}, volume~1, pages
  1073--1083, 2017.

\bibitem[Sethuraman(1994)]{sethuraman1994constructive}
J.~Sethuraman.
\newblock A constructive definition of {Dirichlet} priors.
\newblock \emph{Statistica sinica}, pages 639--650, 1994.

\bibitem[Shaw et~al.(2018{\natexlab{a}})Shaw, Uszkoreit, and
  Vaswani]{shaw2018self}
P.~Shaw, J.~Uszkoreit, and A.~Vaswani.
\newblock Self-attention with relative position representations.
\newblock In \emph{Proceedings of the 2018 Conference of the North American
  Chapter of the Association for Computational Linguistics: Human Language
  Technologies, Volume 2 (Short Papers)}, 2018{\natexlab{a}}.

\bibitem[Shaw et~al.(2018{\natexlab{b}})Shaw, Uszkoreit, and
  Vaswani]{shawSelfAttentionRelativePosition2018}
P.~Shaw, J.~Uszkoreit, and A.~Vaswani.
\newblock Self-{{Attention}} with {{Relative Position Representations}}.
\newblock In \emph{Proceedings of the 2018 {{Conference}} of the {{North
  American Chapter}} of the {{Association}} for {{Computational Linguistics}}:
  {{Human Language Technologies}}, {{Volume}} 2 ({{Short Papers}})}, pages
  464--468. {Association for Computational Linguistics}, 2018{\natexlab{b}}.
\newblock \doi{10.18653/v1/N18-2074}.

\bibitem[Shin(2019)]{spiderschema}
R.~Shin.
\newblock Encoding database schemas with relation-aware self-attention for
  text-to-{SQL} parsers.
\newblock \emph{arXiv preprint arXiv:1906.11790}, 2019.

\bibitem[Sun et~al.(2019)Sun, Zhu, Mou, Xiong, Li, and Zhang]{sun2018grammar}
Z.~Sun, Q.~Zhu, L.~Mou, Y.~Xiong, G.~Li, and L.~Zhang.
\newblock A grammar-based structural {CNN} decoder for code generation.
\newblock In \emph{{AAAI}}, 2019.

\bibitem[Teh and Jordan(2010)]{teh2010hierarchical}
Y.~W. Teh and M.~I. Jordan.
\newblock Hierarchical {Bayesian} nonparametric models with applications.
\newblock \emph{Bayesian nonparametrics}, 1:\penalty0 158--207, 2010.

\bibitem[Vaswani et~al.(2017)Vaswani, Shazeer, Parmar, Uszkoreit, Jones, Gomez,
  Kaiser, and Polosukhin]{vaswaniAttentionAllYou2017}
A.~Vaswani, N.~Shazeer, N.~Parmar, J.~Uszkoreit, L.~Jones, A.~N. Gomez,
  {\L}.~Kaiser, and I.~Polosukhin.
\newblock Attention is all you need.
\newblock In \emph{Advances in {{Neural Information Processing Systems}}},
  pages 5998--6008. {Curran Associates, Inc.}, 2017.

\bibitem[Yin and Neubig(2017)]{yin17acl}
P.~Yin and G.~Neubig.
\newblock A syntactic neural model for general-purpose code generation.
\newblock In \emph{{ACL}}, July 2017.

\bibitem[Yin et~al.(2018{\natexlab{a}})Yin, Deng, Chen, Vasilescu, and
  Neubig]{conala}
P.~Yin, B.~Deng, E.~Chen, B.~Vasilescu, and G.~Neubig.
\newblock Learning to mine aligned code and natural language pairs from
  {StackOverflow}.
\newblock In \emph{International Conference on Mining Software Repositories
  ({MSR})}, pages 476--486. ACM, 2018{\natexlab{a}}.

\bibitem[Yin et~al.(2018{\natexlab{b}})Yin, Zhou, He, and
  Neubig]{yin2018structvae}
P.~Yin, C.~Zhou, J.~He, and G.~Neubig.
\newblock {StructVAE}: Tree-structured latent variable models for
  semi-supervised semantic parsing.
\newblock In \emph{Proceedings of the \nth{56} Annual Meeting of the
  Association for Computational Linguistics ({ACL})}, 2018{\natexlab{b}}.

\bibitem[Yu et~al.(2018)Yu, Zhang, Yang, Yasunaga, Wang, Li, Ma, Li, Yao,
  Roman, Zhang, and Radev]{spider}
T.~Yu, R.~Zhang, K.~Yang, M.~Yasunaga, D.~Wang, Z.~Li, J.~Ma, I.~Li, Q.~Yao,
  S.~Roman, Z.~Zhang, and D.~Radev.
\newblock Spider: A large-scale human-labeled dataset for complex and
  cross-domain semantic parsing and text-to-{SQL} task.
\newblock In \emph{EMNLP}, 2018.
\newblock URL \url{https://yale-lily.github.io/spider}.

\bibitem[Zeiler(2012)]{zeiler2012adadelta}
M.~D. Zeiler.
\newblock Adadelta: an adaptive learning rate method.
\newblock \emph{arXiv preprint arXiv:1212.5701}, 2012.

\end{thebibliography}
